# LLMs and Finetuning: Benchmarking cross-domain performance for hate speech detection

Ahmad Nasir[1], Aadish Sharma[1], Kokil Jaidka[2], and Saifuddin Ahmed[3]

[1] Indian Institute of Technology Delhi, India
{nasirahmadiitd, aadisharma.iitd}@gmail.com
[2] Communications and New Media, National University of Singapore, Singapore
jaidka@nus.edu.sg
[3] Wee Kim Wee School of Communcation and Information, Nanyang Technological University, Singapore
[sahmed@ntu.edu.sg]

**Abstract.** In the evolving landscape of online communication, hate speech detection remains a formidable challenge, further compounded by the diversity of digital platforms. This study investigates the effectiveness and adaptability of pre-trained and fine-tuned Large Language Models (LLMs) in identifying hate speech, to address two central questions: (1) To what extent does the model performance depend on the fine-tuning and training parameters?, (2) To what extent do models generalize to cross-domain hate speech detection? and (3) What are the specific features of the datasets or models that influence the generalization potential? The experiment shows that LLMs offer a huge advantage over the state-of-the-art even without pretraining. Ordinary least squares analyses suggest that the advantage of training with fine-grained hate speech labels is washed away with the increase in dataset size. While our research demonstrates the potential of large language models (LLMs) for hate speech detection, several limitations remain, particularly regarding the validity and the reproducibility of the results. We conclude with an exhaustive discussion of the challenges we faced in our experimentation and offer recommended best practices for future scholars designing benchmarking experiments of this kind.

## 1 Introduction

The widespread use of the internet, especially social media, has amplified online hate speech, with varying regulations across regions like the US and the EU [14]. While platforms like Facebook employ automated systems to remove potential hate content, certain severe forms of hate speech, such as threats or racism, require legal action. As these platforms become increasingly prevalent, effectively identifying and mitigating such harmful content becomes paramount. Given the high costs and psychological toll of human moderation, there is a pressing need for automated solutions.

However, hate speech detection is a challenging problem where a single model may not generalize to different scenarios. For instance, differentiating between hateful, spam, abusive, and profane content requires a deep understanding of the underlying intent and the context in which the words are used [30,18,39]. Large language models (LLMs) may offer a promising solution for high-precision hate speech detection, as they are often trained on extensive, multilingual and multi-source text data for various text classification tasks [29,31,4]. However, in this case, benchmarking black-box large language models (LLMs) remains a significant challenge. This is because the opacity surrounding how LLMs are trained – specifically, what their training data comprised, and what specific tasks they were trained on – poses a considerable barrier to their effective utilization. Without this knowledge, determining the most appropriate LLM for a task such as hate speech detection becomes an increasingly difficult guessing game, in part due to the proliferation of LLMs that successively outperform each other on a standard set of tasks [4]. The lack of transparency also raises concerns about their generalizability to real-world scenarios.

From an application standpoint, a second research gap emerges regarding whether fine-tuning would improve the model performance for a popular task such as hate speech detection, as there is no prior evidence to suggest that it consistently enhances performance across various domains. A third research gap lies in whether models, if finetuned and/or pre-trained, would generalize easily to new domains and what factors are predictive of generalizability. Prior research has referred to the importance of dataset size, label similarity [15,28], coverage [22], and the number of authors [2], yet they do not offer an exhaustive analysis of the effects of dataset characteristics on model generalizability. Furthermore, although cross-domain classification studies of hate speech detection have been conducted in the past [10], similar benchmarks are unavailable in the context of LLMs.

To address these research gaps, in this study, we evaluate and benchmark various standard and fine-tuned large language models (LLMs) for detecting hate speech in different domains and contexts. Our analytical framework serves as a benchmarking tool, setting a standard for future research in hate speech detection using LLMs, ensuring consistent and comparable results. Our work addresses the following research questions:

1. To what extent does the model performance depend on the fine-tuning and training parameters?
2. To what extent do models generalize to cross-domain hate speech detection?
3. What are the specific features of the datasets or models that influence the generalization potential?

| Dataset | Instances | Instances after preprocessing | Source | Positive (1) | Negative (0) |
|---|---|---|---|---|---|
| Gab | 33,776 | 32,010 | Gab | 14,614 | 17,396 |
| Reddit | 22,334 | 16,982 | Reddit | 5,255 | 11,727 |
| OLID | 14,200 | 9,914 | Twitter | 1,074 | 8,840 |
| $SOLID_{Extended}$ | 3340 | 3,337 | Twitter | 347 | 2,990 |
| $SOLID_{SemEval}$ | 2997 | 2,995 | Twitter | 188 | 2,807 |
| HASOC (English) | 5,983 | 4,734 | Twitter and Facebook | 1,143 | 3,591 |
| ICWSM | 5,143 | 3,221 | YouTube and Facebook | 2,364 | 857 |
| Restricted | 100,000 | 58,812 | Twitter | 4,963 | 53,849 |
| Toraman (English) | 100,000 | 89,236 | Twitter | 1,617 | 87,619 |

**Table 1.** The datasets used in our experiments. 1 = "Hateful" and 0 = "Non-Hateful"

## 2 Background

Cross-domain classification is the generalization of a classification model trained on a particular type of dataset and used to classify datasets that the model has not seen while training. There has recently been a spurt in studies that offer a cross-domain setting for hate speech detection [33,34,6,3,36,13,15,28,26,9,1,37].

The studies by [33] and [35] collapsed the data into hate speech and not hate speech and tested their models on out-of-domain data, reporting poor generalizability. [3] applied a different approach to generalize models to new datasets. Given unlabeled data from multiple sources, they used a semantic distance measure to distinguish hate speech from others. The authors found that this approach outperformed models that were trained on in-domain labeled data and could generalize well. The study by [36] was able to generalize to new data by enriching their training data with a third dataset and then using multi-task learning. [15] used 9 different datasets - W&H [35], Waseem [33], TRAC [17], Kol [16], Gao [12], Kaggle [5], Wul1, Wul2 and Wul3 [38] in training their hate speech models. First, they categorized all datasets into positive (abusive) and negative (non-abusive), then trained each of the datasets using support vector machines and tested on 8 other datasets, they observed 50% reduction in F1 score on out-of-domain data. Next, they used transfer learning to augment the original domain with the information from a different domain. Their approach resulted in improvement in 6 out of 9 cross-domain datasets, and the authors concluded that achieving strong performance in classifying the target dataset requires including at least some training data from that specific dataset.

However, the findings from [28] and [10] offer a different perspective. Unlike [36] and [15], they showed generalization can be achieved without augmenting out-of-domain data. [28] merged datasets into binary categories and trained state-of-the-art BERT-based models. They reported that in-domain results were significantly better than other state-of-the-art models just by fine-tuning. They also extrapolated that the model would perform better on out-of-domain data if the datasets were of a similar type. Furthermore, their findings suggest that datasets with larger percentages of positive samples tend to generalize better than datasets with fewer positive samples, in particular when tested against dissimilar datasets. [10] used 9 publicly available datasets and collapsed them into binary categories and were able to train BERT, ALBERT, SVM, and fastText. They observed that in most of the cases transformer models generalize better. For generalization, in-domain performance is also important.

In another study, [2] bring up an additional characteristic that should be taken into account in the context of cross-dataset hate speech classification, namely the number of authors of the material captured in a dataset; however, the authors report an F1 score of 0.54, suggesting that dataset size and author complexity may increase the difficulty of the task.

Recent work has explored similar concerns. For example, [1] provide a large-scale empirical evaluation across multiple datasets and show that cross-dataset performance is often overestimated due to insufficient methodological rigor. Likewise, [37] introduce a novel framework to audit model disagreement with humans on what counts as offensive, surfacing errors in both labeling and model behavior. Complementing these, [9] offer strategies for generalizing offensive language identification models across domains using low-resource fine-tuning. Other studies have introduced novel hate speech datasets drawn from specific sociocultural contexts, such as online chess communities [27], or redefined what counts as “extreme speech” in consultation with affected communities [20].

In this work, we address some of the pressing gaps that remain in prior work, such as the need to relate the advances in natural language processing research to the body of prior work on cross-domain studies of hate speech detection. Previous studies have achieved the best performance using different approaches, ranging from SVMs to deep learning models (e.g., [13]). However, in recent years, transformer models like BERT have outperformed most others. Transfer learning has also been shown to improve BERT performance [21]. Our work benchmarks the performance of LLMs on hate speech detection tasks for many standard datasets against these prior state-of-the-art results, while also setting an agenda for future research.

## 3 Approach

We aim to understand the robustness and adaptability of our models in identifying hate speech across varied contexts and sources. Therefore, we have reported the efficacy of LLMs in detecting hate speech as in-domain performance, as well as across different domains (cross-domain), both pre- and post-fine-tuning.

### 3.1 Model Setup

We have illustrated our model pipline in Figure 1 and described it in the following paragraphs.

As our goal is to train large language models to classify an input text into two classes, i.e. Neutral or Hateful, we attempted a few approaches in building the classification pipeline. First, we used the LLM to generate textual output and used techniques like keyword extraction to classify it into two classes. Second, we used the LLM to generate textual output, get the BERT embeddings of the generated text, and build a classifier using the BERT embeddings to classify it into the two classes. Finally, we used the LLM to get the last hidden layer's embeddings and build a classifier using the hidden embeddings to classify it into the two classes.

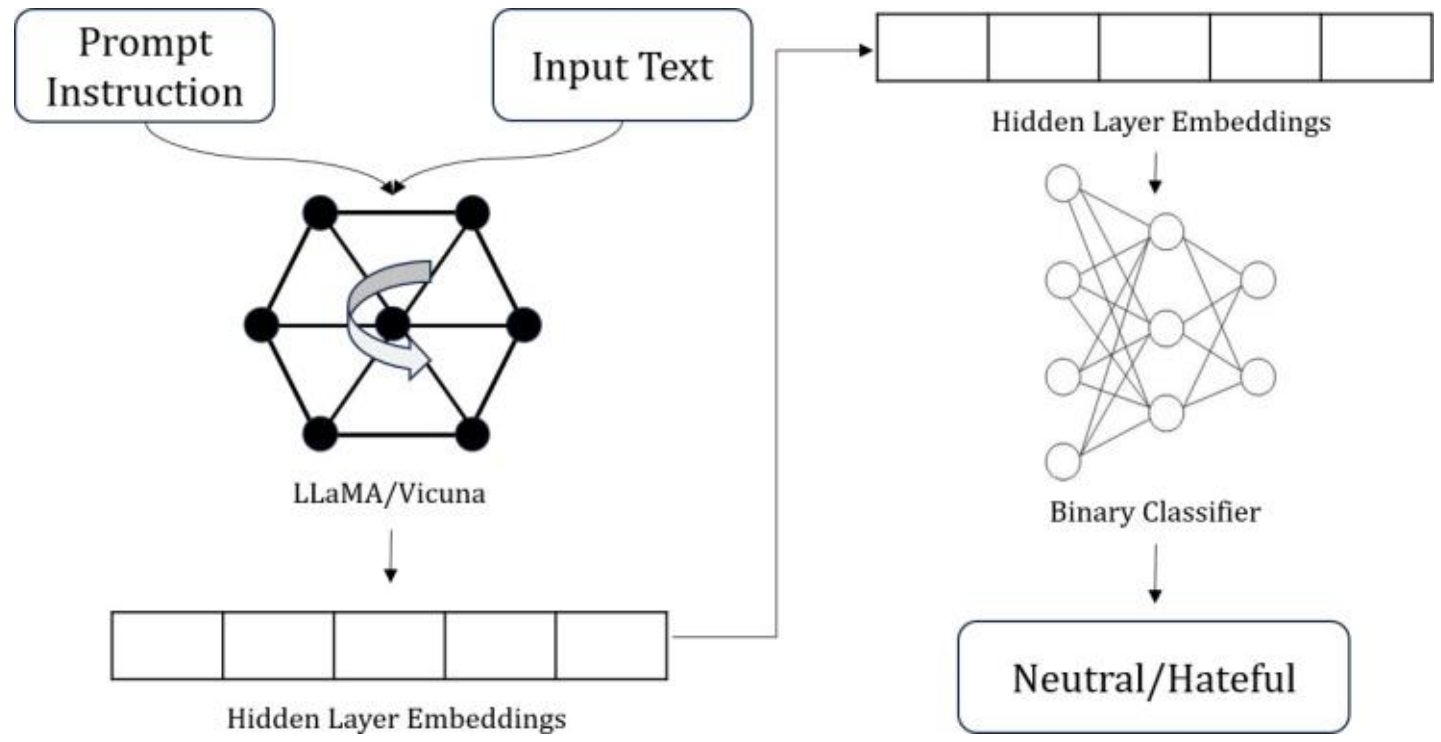

**Fig. 1.** Model Pipeline

We had very limited success with the first two techniques, perhaps because we were trying to fit the model's textual output into the two classes, which contains many extra output words apart from the correct classification. Thus, we chose the third technique to build our framework. The framework comprises a classification pipeline using a Backbone LLM to get text embeddings. The LLM may or may not be fine-tuned on a sub-sample of the hate speech dataset, depending on the designated parameters of the pipeline. Subsequently, we use the embeddings to classify it as Neutral or Hateful with the help of a Binary Classifier. First, we provide an input text, converted into the following instruction format <*Instruction, Input, Output*>. Next, the instruction format input is fed to a Backbone LLM model, and the LLM's last hidden layer is extracted. Finally, the last hidden layer is fed to a Binary Classification model, which outputs the predicted class.

**Backbone Model** The LLM Backbone can be a Large Language Model, such as LLaMA [31], Alpaca [29], Vicuna [4] etc. In our paper, we have used the following 4 Backbone Models:

- LLaMA-7B[4]: LLaMA is open source foundation model released by Meta AI. It has been trained on publicly accessible data. It is based on the transformer architecture proposed by [32] and comprises of 7 billion parameters.
- LLaMA-7B-finetuned: We instruction fine-tune the base LLaMA-7B model on a sub-sample of our hate-speech data using QLoRA [7] and PEFT [19] which is a cost-effective technique for low resource training.

[4] https://huggingface.co/decapoda-research/llama-7b-hf

- Vicuna-7B:[5] Vicuna-7B is obtained by fine-tuning LLaMA-7B base model using approximately 70K user-shared conversations gathered from ShareGPT.com
- Vicuna-7B-finetuned: Similar to the fine-tuning of base LLaMA-7B, we have used instruction-based fine-tuning to fine-tune the Vicuna-7B base model on a sub-sample of our hate-speech dataset.

The reason for our choice of the above two base models, viz. LLaMA [31] and Vicuna [4] is that while LLaMA is a foundation model, which has not been fine-tuned on any specific task, Vicuna is built on top of Alpaca [29], and Alpaca has been fine-tuned to follow instructions. Thus, we wanted to compare the performance of foundation and instruction-tuned models.

**Binary Classifier** The Binary Classifier is trained on the last hidden layer embeddings of the Backbone LLMs. We convert all the training samples to the *Alpaca Instruction Format*, and pass them through the Backbone LLMs, and dump the last hidden layer weights. Then, we use these 4096-dimensional embeddings as input to the classifier and 0/1 as labels for Neutral/Hateful cases, respectively, to train the classifier.

### 3.2 LLM Fine-tuning and Classifier Training

**LLM Fine-tuning** We fine-tuned both of the base models, i.e. LLaMA and Vicuna on a sub-sample of our hate-speech dataset, using QLoRA [7], an extremely cost-effective and affordable training technique. By fine-tuning the base models on our hate-speech dataset, we hope to help the model learn various nuances of the actual examples of hateful texts.

We used 1000 examples from each of the 9 datasets and converted them in the Alpaca Instruction Format <*Instruction, Input, Output*> to do the instruction fine-tuning. We used *"Detect if the text is hateful or neutral"* as the instruction, the content of the input text as the input, and *HATEFUL or NEUTRAL* as the output. Making use of fine-tuning scripts[6,7], we ran the fine-tuning for 1 epoch with batch size 128. This provided us with two more Backbone LLMs, i.e. *LLaMA-7B-finetuned*, and *Vicuna-7B-finetuned*.

**Classifier Training** For a given dataset, the Binary Classifier is trained on the last hidden layer embeddings of the Backbone LLMs. We take 2000 training examples from the given dataset, and convert them to *Alpaca Instruction Format*, and pass them through one of the four Backbone LLMs, and dump the last hidden layer weights. Then, we use these dumped embeddings as input to the classifier and 0/1 as labels for Neutral/Hateful cases, respectively, to train the classifier. Thus, for each dataset, we end up with 4 Binary Classifiers, one for each Backbone LLM. The classifier is trained for 50 epochs, with early stopping, and in most cases, training ends at around 15-20 epochs.

[5] https://huggingface.co/lmsys/vicuna-7b-v1.5

[6] https://github.com/tloen/alpaca-lora

[7] https://github.com/artidoro/qlora

As a result of Binary Classification training, we have 4 classifiers for each of the 9 datasets, and we use these classifiers to evaluate the self-domain and cross-domain performance on the test sets (end-domains).

Also, when we are training the classifiers on the fine-tuned models, we observe that using a model fine-tuned on an enriched dataset (9000 samples, 1000 from each dataset) performs significantly better than non-enriched dataset (only 1000 samples from the respective dataset), thus we decide to use the models fine-tuned on the enriched datasets to get the weight dumps to train the classifiers.

# 4 Experimental setup

In the *In-Domain Performance* section, we evaluated various Backbone LLMs on their performance on hate speech detection across nine datasets reported in Table 1. Therefore, each LLM is paired with 9 classifiers tailored to a specific dataset. These classifiers are trained on 2,000 samples from their respective datasets and subsequently tested on the entirety of the same dataset. The results comparing the best model performance on the F1 scores are reported in Table 2.

In the *Cross-Domain Performance* evaluation, we explored the adaptability of these models. Each backbone LLM is associated with 8 out-of-domain classifiers from different domains for every end-domain dataset. Though trained on 2,000 samples from their datasets, these classifiers are evaluated on all samples from a different end-domain dataset to obtain cross-domain F1 scores. The results are detailed in Table 3.

Finally, for the evaluation of factors predicting *Model Generalizability*,we consider the predictive performance as a multivariate optimization problem dependent on the source similarity, source of origin, label distribution, dataset size, base model, and other considerations.

Therefore, our experiments denote a $5(source) \times 2(basemodel) \times 2(finetuning)$ ablation design, with additional covariates, such as training dataset size and in-domain performance. We then conducted a multiple linear regression over these parameters to identify which of these factors played a significant role in explaining model performance in a cross-domain setting.

## 4.1 Datasets

The Hate Speech detection task was formulated as a binary classification problem of whether a text is hateful or non-hateful, on nine publicly available datasets, with label distributions reported in Table 1.

- Reddit and Gab [23]: These datasets comprise of 22,334 and 33,776 points respectively and the texts are categorised as hateful and non-hateful. Reddit dataset is more imbalanced than Gab. For testing and training we have kept the same ratios as present in the data.

- HASOC [18]: This dataset has 5 different columns HateSpeech, Offensive, Profanity, Non-Hate-Offensive and Hate-Offensive. Our study has used Hate-Speech and Non-Hate-Offensive columns from the English dataset.
- Toraman [30]: This dataset has 100,000 tweets; 20k tweets are taken from each different domain - religion, sports, gender, race and politics, with columns as hate, offensive and normal.
- ICWSM [25]: This is comments data on videos posted by social media news on YouTube and Facebook. The comments are divided into hateful and normal and further divided into 13 main categories and 16 subcategories.
- OLID [39], SOLID SemEval and SOLID Extended [24]: The datasets are hierarchically divided into three tasks - Offensive Language Detection, Categorization of Offensive Language, and Offensive Language Target Identification. We have used the first task to get the neutral examples and the third task to get the hateful examples.
- Restricted [11]: Comprising of 100,000 annotated data points categorized into four distinct classes: normal, hateful, spam, and abusive. We have segregated normal and hateful classes as per the need of our study.

## 5 Results & Analysis

The following paragraphs summarize our key results, while the detailed model performance is reported in the online supplement.[8]

### 5.1 In-Domain Performance

In answer to RQ1, we observe that using LLMs, even without fine-tuning, significantly improves the self-domain as well as the cross-domain performance on the datasets, as compared to the previously available best models, as reported in Table 2

Table 2 provides the best results per dataset per model variant. For Gab, the F1-score of the best model (CNN) was 0.896, while that of our best model is 0.9942, which is an almost 10% increase (0.0982). For Reddit, the F1-score of the best model (RNN) was 0.77, which increased to 0.9601, resulting in a substantive 19% (0.1901) increase. For the Toraman, the F1-score of the best model (Megatron) was 0.830, which increased to 0.9851 for our best model, resulting in 15% (0.1551) increase in the F1-score. For ICWSM, the F1-score of the best model (SVM) was 0.96, which increased to 0.9978 for our best model.

For datasets like OLID, SOLID_SemEval, SOLID_Extended and HASOC, there isn't a direct comparison, as their dataset is in a hierarchical format while ours is in a binary classification format, but even for the first task, our best model's performance significantly outperforms the baselines, as illustrated in the Self-Domain Performance.

As per Table 2, in about half of the cases the base model for LLaMA outperformed the fine-tuned model. Note the very high scores on the three Solid

[8] https://anonymous.4open.science/r/colm$_{r}esults - E$888/

datasets, which leads us to suspect that they are pre-included in the models' pretraining corpora.

By instruction fine-tuning the base LLaMA-7B model, we observe that the average F1-score of LLaMA-7B-finetuned model slightly decreases (0.00304) as compared to the LLaMA-7B model while at the same time, by instruction fine-tuning Vicuna-7B, the average F1-score of Vicuna-7B-finetuned increases (0.001319) as compared to the Vicuna-7B model. There is a significant improvement in the F1-score of Toraman, Reddit and OLID.

| Dataset | LLaMA Variants | | Vicuna Variants | |
|---|---|---|---|---|
| | Best-Performing Variant (F1) | F1 | Best-Performing Variant (F1) | F1 |
| Gab | Training | 0.994 | Training | 0.994 |
| HASOC | Finetuning + Training | 0.997 | Finetuning + Training | 0.997 |
| ICWSM | Finetuning + Training | 0.998 | Training | 0.998 |
| Reddit | Finetuning + Training | 0.959 | Finetuning + Training | 0.960 |
| Restricted | Training | 0.994 | Training | 0.998 |
| Toraman | Training | 0.978 | Finetuning + Training | 0.985 |
| Solid Extended | Training | 1 | Training | 1 |
| OLID | Training | 0.998 | Finetuning + Training | 0.999 |
| Solid SemEval | Training | 1 | Finetuning + Training | 0.997 |

**Table 2.** The best-performing LLaMA and Vicuna variants for in-domain hate speech detection.

| Test Dataset | Best Performing Model | Training Dataset | F1 |
|---|---|---|---|
| Gab | Vicuna | Reddit | 0.992 |
| Reddit | LLaMA | Gab | 0.952 |
| HASOC | Vicuna | Gab | 0.997 |
| ICWSM | LLaMA | HASOC | 0.995 |
| Restricted | LLaMA | Gab | 0.998 |
| Toraman | Vicuna | Gab, OLID | 0.993 |
| OLID | LLaMA | Gab, Reddit | 0.999 |
| Solid Extended | LLaMA | Gab, HASOC | 1 |
| Solid SemEval | LLaMA | Gab, HASOC | 1 |

**Table 3.** The best-performing models for cross-domain hate speech detection.

## 5.2 Cross-Domain Performance

In answer to RQ2, Table 3 and Figure 2 reports the best-performing fine-tuned models for cross-domain hate speech detection, with the second column denoting the best or fine-tuned models for each validation dataset in the third column. In general, the Table suggests that models fine-tuned on the Gab dataset had the best cross-domain generalizability, except in the case of ICWSM (sourced from Twitter), where a model fine-tuned on HASOC, also sourced from Twitter, performed the best. Once again, the OLID, SOLID-Extended,and SOLID SemEval datasets proved to be less useful benchmarks as suggested by the extremely high model accuracies. We expect that this occurred because they may have been used for pretraining the LLMs we chose, and have little to do with the dataset we fine-tuned on.

**Fig. 2.** Cross-domain predictive performance for hate speech detection

The findings suggest that instruction fine-tuning Vicuna-7B model has resulted in the increase of the cross-domain performance for 7 out of the 9 datasets (Gab, ICWSM, OLID, Reddit, Restricted, SOLID_Extended and SOLID_SemEval) as compared to the base Vicuna-7B model. Similarly, instruction fine-tuning LLaMA-7B has resulted in the increase of the cross-domain performance for 5 out of 9 datasets (HASOC, ICWSM, Restricted, SOLID_SemEval and Toraman) as compared to the base LLaMA model. Instances of false negatives in the fine-tuned Vicuna backbone trained on Gab (typically the best-performing cross-domain setup in our findings) and tested on Reddit, and vice-versa, are reported in the Appendix in Table 4. Many of the instances of the Gab-trained model tested on Reddit fail on the subtle instances of hate speech, while the Reddit-trained model appear to fail on both subtle and explicitly hateful instances.

### 5.3 Error Analysis

Table 4 reports some of the examples where some of the best-performing cross-domain finetuned Vicuna models (as per Table 3) failed to correctly classify the input (false negatives).

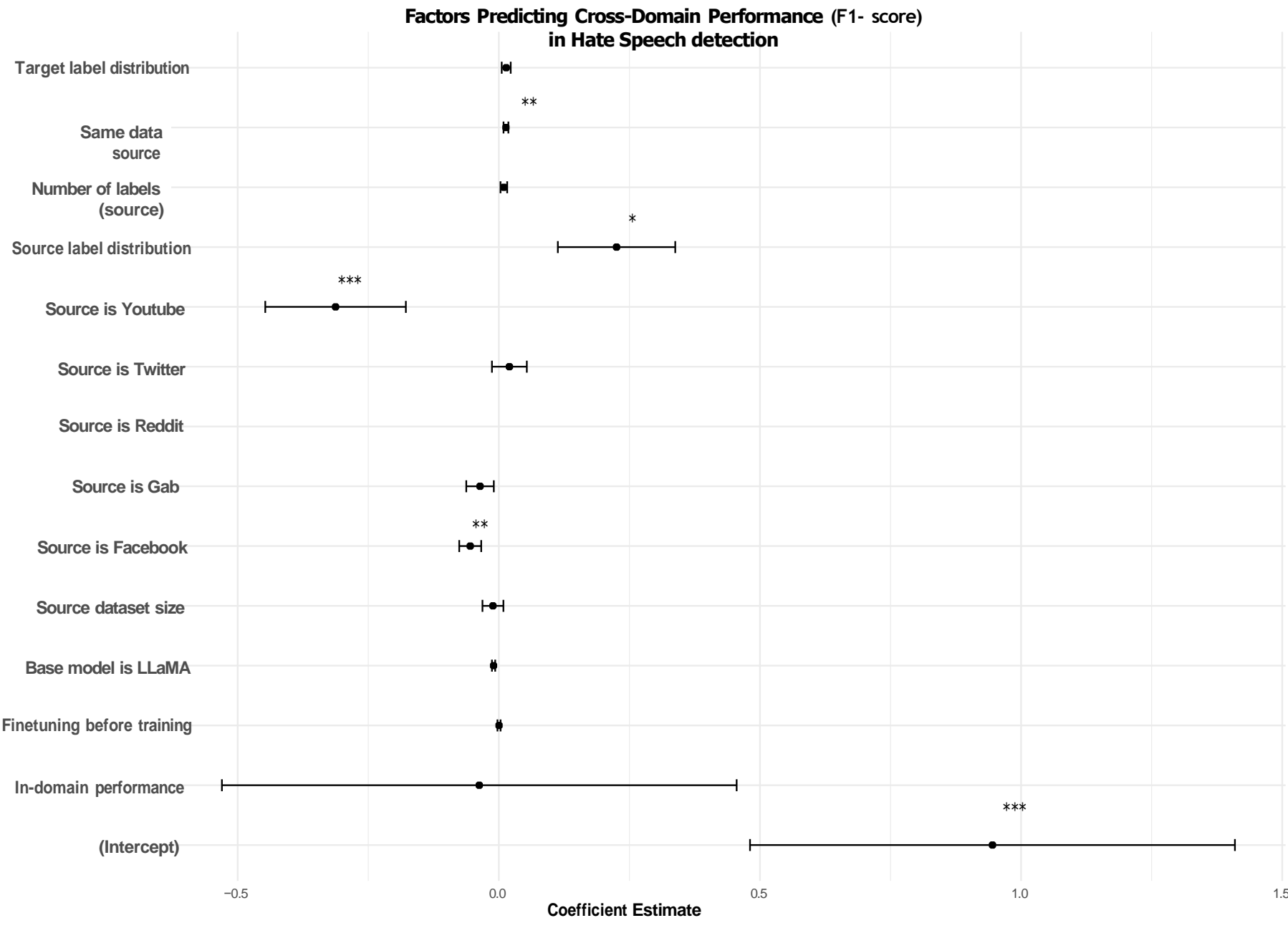

**Fig. 3.** Model generalizability measured through regressing all model performance on training and testing parameters

In general, we observed that false negatives occur when the model struggles to detect implicit toxicity or subtle offensive language. This is especially evident in cases where the hate speech is masked by sarcasm or indirect phrasing (e.g., "A chinaman and a commie teaming up?"). In particular, long and convoluted sentences that bury toxic elements are often missed (e.g., "The flag is helpless to stop you..."). Similarly, the classifier tends to overlook offensive slurs that are casually integrated into discourse (e.g., "You argue like a child, although..."). Moreover, slurs and offensive terms that are normalized in certain online communities seem to evade detection more frequently (e.g., "Get on the Front line C***!!"). On platforms like Gab or Reddit, where language norms differ, the model often fails to adapt (e.g., "Rick, the couple people I know..."). For instance, terms used casually in one platform might not trigger the model's filters on another (e.g., "C*** hair away from achieving the same..."). The model also struggles with offensive terms that appear at the end of otherwise neutral statements (e.g., "It only shows you're a..."). While certain phrases may seem contextually benign, their deeper connotations may go unnoticed (e.g., "Kanye saw through Taylor Tw** Swift..."). Lastly, the classifier's reliance on explicit markers of toxicity may limit its ability to capture more nuanced expressions of hate.

**Classifier with a fine-tuned Vicuna backbone LLM, trained on Reddit, tested on Gab**

- A chinaman and a commie teaming up? The great proletarian cultural revolution is upon us!
- I think Kanye saw through Taylor Twat Swift before any of us saw who she really is. I question no more about why Kanye started cracking on her by taking the microphone away.
- Politicians misspeak all the time Yes I'm sure Kasich, a RINO and Trump hater, conspired to put McCain to death and knew about it advance. When McCain was a RINO and loved by other RINOs. ????? You are dumber than a black African.
- Twat, he loves them because they finance him. Book deals are a dead give away.
- Give it to a soldier who has defend it , then try to burn it . Show me what a badass you really are. The flag is helpless to stop you or fight back . Completely unnecessary act , its proves nothing and accomplishes nothing. It only shows your a ignorant weak faggot . Like kicking a puppy „ , it shows how weak and worthless you are
- Trump may not be our guy but he proved in the primaries that it's possible to mount a successful primary challenge against multiple deep state insiders and media favorites. Ron Paul was a cunt hair away from achieving the same, himself.
- As I said previously, you argue like a child, although with additional feedback I'd argue a retarded one at best.
- Get on the Front line Cunt !! Quit mouthin off & Get out there !!
- Astronaughts should stick to what they know best... getting out there in space...perhaps the silly cunt should take his helmet off next time he's having a spacewalk. Fuck me these lefty scum give me the literal shits...and some of these useless cunts think that having fucked the good ole USA up the ass with their liberal crapola they can move to New Zealand

**Classifier with a fine-tuned Vicuna backbone LLM, trained on Gab, tested on Reddit**

- I'll agree on Rick, the couple people I know who say he's one of their favorite characters are insufferable twats.
- Look a historical revisionist, how cute. Go back to your gender studies echo chamber and plot on how the world would be perfect if men simply didn't exist.
- Oh wow we got a white knight here that goes on PPD and starts talking shit about commenters. Why don't you fly to England, put on your white knight armor, and defend all the fat worthless alimony money stealing fatties.
- An american will start yelling or disparaging. Verbal sparring is fun and really interesting to watch... but a bumbling piece of white trash is neither.

**Table 4.** Examples of false negatives in cross-domain hate speech detection.

## 5.4 Factors determining model generalizability

The ordinary least squares analyses reported in Figure 3 suggest that to generalize well, training dataset size and in-domain performance are secondary to label imbalance. Simple, small-sample, pre-training of Vicuna-based language models rather than fine-tuning significantly impact model performance. The effects of having fine-grained labels matter primarily when training datasets are small in size; however, these effects are washed away for larger training datasets. We see that fine-tuning LLMs matters little, and researchers may simply pre-train their chosen language models on a small labeled sample, preferably from the same source, in order to get the best possible results. Fine-tuning or training on Facebook or YouTube data appears to adversely impact model performance, more so in the case of YouTube than Facebook.

# 6 Discussion

Our findings suggest that cross-domain fine-tuning, even on a single epoch, is nearly always beneficial for greater precision on a known target dataset. Although newer models have emerged, these findings remain highly relevant, as the principles of enhancing model precision through targeted fine-tuning and leveraging label diversity apply to current and future models. Therefore, our results raise questions about the need to fine-tune large language models (LLMs) for text classification tasks, such as hate speech detection. While models like Vicuna-7B, when fine-tuned, consistently outperform their base versions across various datasets, this is not the case with LLaMA-7B.

Furthermore, our findings provide a complementary perspective to the insights from [28] and [10]. While their findings suggested that training data from a similar platform or a model with high in-domain performance was a critical factor for achieving effective generalization in hate speech detection, our analysis suggests that label distribution offers a key indicator of effective fine-tuning for LLMs.

## 6.1 Limitations

While our research demonstrates the potential of large language models (LLMs) for hate speech detection, several limitations remain, particularly regarding the validity and the reproducibility of the results. We offer these in the interest of transparency, as points of discussion, and as recommended best practices for future scholars designing benchmarking experiments of this kind.

First, hate speech detection is highly context-dependent, and the same model may not generalize well across different sociocultural settings or types of hate speech. Differentiating between hateful, abusive, spam, or profane content requires a deep understanding of the language and intent behind it, which current LLMs may not fully capture [30].

Second, we acknowledge that aspects of our experimental design and implementation setup introduced severe limitations, several of which were raised by reviewers. Our evaluation framework—where each LLM is paired with nine classifiers, each trained on 2,000 examples from a single dataset and tested on the same dataset—may inadvertently reinforce dataset-specific artifacts. While this approach standardizes training size across datasets, it also risks overfitting to dataset idiosyncrasies rather than surfacing generalization capabilities. Reviewers were concerned about the use of overlapping training and test sets, which risked test-train contamination, and the implausibly high F1 scores reported across all datasets—including subjective hate speech datasets known to suffer from annotation noise.

Thirdly, reviewers were concerned that the results looked “too good to be true” in some cases for a task like offensive language detection. Many approaches achieved almost perfect F1 scores. The opacity of LLMs regarding their training data pose a significant challenge for us to diagnose whether this was an issue in our benchmarking experiments or in selecting models for specific tasks like hate speech detection [4]. Without transparency regarding the datasets and tasks used for pre-training LL<s, it is challenging to determine a model's suitability for benchmarking experiments as well as real-world applications.

Lastly, while we observed cross-domain fine-tuning consistently improving performance, the lack of labeled data for certain hate speech subtypes and label imbalances across datasets may limit the generalizability of our findings. This is particularly concerning for underrepresented or emergent forms of hate speech, which may not be adequately captured by models trained on canonical datasets like SOLID or Semeval.

Despite these valid concerns, we would like to defend our design, which was chosen based on several principles. Our goal was to simulate low-resource, high-variance environments typical of real-world hate speech detection settings, where only a small amount of labeled data is available per domain. Fixing the training size to 2,000 samples allowed us to control for dataset size effects while testing on the entire dataset helped us capture model behavior on rare or ambiguous cases. We prioritized consistency and comparability across all nine datasets, striking a balance between practical constraints and analytical scope. In practice, downstream applications often involve inference on previously unseen data distributions, and such per-dataset classifiers may underperform in such settings.

Nonetheless, a major concern we would like to highlight is that our replication attempts were hindered by model versioning and package incompatibilities, which prevented us from reproducing our own results within a short span of three months. At the time of our experiments, the model and tokenizer versions we used were stable and well-supported; it was only with subsequent updates to LLaMA, Vicuna, and core HuggingFace libraries that incompatibilities arose, which we regret not anticipating or isolating earlier.

## 7 Conclusion

Our contribution is a research design to benchmark cross-domain performance of hate speech detection models. We demonstrate substantive improvements over the state-of-the-art hate speech detection models, often with a single epoch of fine-tuning. The first key takeaway of our work is that generalization varies from model to model and that, in general, label imbalance affects model performance in cross-dataset applications. The second takeaway is that for out-of-domain prediction on a given end-domain dataset, there are specific advantages for fine-tuned models trained on small datasets with fine-grained hate speech labels.

In light of the limitations and community feedback, we propose the following recommendations for future work:

- Use clearly separated train, validation, and test sets with no content overlap; document all sampling and split procedures.
- Avoid evaluating on full datasets after training on partial subsets; instead, use dedicated held-out sets or k-fold cross-validation.

– Track and version all dependencies, including model weights, tokenizers, and environments, using tools such as Docker or conda-lock.
– Include comparisons with standard baselines (e.g., CNNs, RNNs, or instruction-tuned LLMs) for interpretability and external validity.
– Report the full range of performance metrics across all tested configurations, not only the top-performing model.
– Use the term cross-dataset unless there is demonstrable variation in domain characteristics such as platform, language variety, or sociocultural context.
– Acknowledge and analyze the definitional and annotation inconsistencies across hate speech datasets, particularly when interpreting cross-dataset generalization results.

These best practices are necessary for improving the interpretability, reproducibility, and credibility of LLM-based hate speech detection studies.

**Ethical considerations:** We caution against relying exclusively on AI-generated predictions of hate speech for high-stakes or real-world applications. Hate speech is inherently subjective and highly context-dependent; even accurate models at the aggregate level may fail to account for cultural nuance, sarcasm, or reclaimed language at the individual level. These risks are amplified when models trained in one sociocultural context are applied to another, where norms and linguistic cues differ [8]. Furthermore, the opacity of large language models and the lack of transparency in their pretraining data make it difficult to assess how sociocultural priors are encoded or reinforced.

To mitigate these risks, we emphasize the need for continual in-context evaluation, supported by human oversight and participatory annotation practices that reflect the values and expectations of affected communities. Transparency, context-awareness, and sustained monitoring are essential for the responsible deployment of LLMs in hate speech detection.

**Acknowledgments:** This research was supported by the Ministry of Education, Singapore, through its MOE AcRF Tier 3 Grant (MOE- MOET32022-0001) and MOE Tier 1 Grant (WBS A-8000231-01-00). We also extend our gratitude to the anonymous reviewers of the Conference on Language Modeling (COLM) 2024 and the International Conference on Computational Linguistics (COLING) 2025 for their invaluable feedback.

# A Baseline results

Table 5 reports the state-of-the-art in prior work for each of the nine datasets under study, for the task of hate speech detection.

| Dataset | Best Performing Model | Metric | Score |
|---|---|---|---|
| Gab | CNN | F1 | 0.896 |
| Reddit | RNN | F1 | 0.775 |
| Toraman | Megatron | F1 | 0.830 |
| ICWSM | SVM | F1 | 0.960 |
| OLID (Task A) | CNN | F1 | 0.900 |
| OLID (Task C) | CNN | F1 | 0.670 |
| SOLID SemEval (Task A) | BERT | Macro-F1 | 0.923 |
| SOLID SemEval (Task C) | BERT | Macro-F1 | 0.645 |
| HASOC (Task A) | LSTM | Weighted F1 | 0.839 |
| HASOC (Task C) | LSTM | Weighted F1 | 0.820 |

**Table 5.** Baseline results of the best model for each dataset

# B Additional results

The comprehensive set of results for the 36 classifiers evaluated for RQ1 and the 288 classifiers evaluated for RQ2 are reported in < https://anonymous.4open.science/r/colm_results-E888/ >.